\documentclass{article}
\usepackage{spconf,amsmath,graphicx}
\usepackage{amsfonts}
\usepackage{appendix}
\usepackage{color}

\usepackage{soul}

\newcommand{\EE}{{\mathbb E}}
\newcommand{\wb}{{\bf w}}
\newcommand{\xb}{{\bf x}}

\newcommand{\mub}{\boldsymbol\mu}

\newcommand{\Sigmab}{\boldsymbol\Sigma}

\title{A probabilistic Least-Mean-Squares Filter} 

\twoauthors
  {Jes\'us Fern\'andez-Bes, V\'ictor Elvira\thanks{The work of Jes\'us Fern\'andez-Bes was supported by the Spanish MECD FPU program and MICINN project PRI-PIBIN-2011-1266. The work of V\'ictor Elvira was supported by the Spanish Government (MINECO) under project OTOSiS (TEC2013-41718-R), the Community of Madrid Government under project AGES (S2010/BMD-2422), and the European Union's 7th FP through the Marie Curie ITN MLPM2012 (Grant No. 316861).}}
   {\small
   Department of Signal Theory and Communications\\
   \small Universidad Carlos III de Madrid\\
   \small Legan\'es (Spain)}
  {Steven Van Vaerenbergh\thanks{The work of Steven Van Vaerenbergh was supported by the Spanish Government (MINECO) under project RACHEL (TEC2013-47141-C4-3-R).}}
   {\small Department of Communications Engineering \\
   \small University  of Cantabria\\
   \small Santander (Spain)}
   
\begin{document}
\maketitle

\begin{abstract}%

We introduce a probabilistic approach to the LMS filter. By means of an efficient approximation, this approach provides an adaptable step-size LMS algorithm together with a measure of uncertainty about the estimation. In addition, the proposed approximation preserves the linear complexity of the standard LMS. Numerical results show the improved performance of the algorithm with respect to standard LMS and state-of-the-art algorithms with similar complexity. The goal of this work, therefore, is to open the door to bring some more Bayesian machine learning techniques to adaptive filtering.
\end{abstract}


\section{Introduction}
\label{sec:introduction}

Probabilistic models have proven to be very useful in a lot of applications in signal processing where signal estimation is needed \cite{rabiner1989tutorial,arulampalam2002tutorial,ji2008bayesian}. Some of their advantages are that 1) they force the designer to specify all the assumptions of the model, 2) they provide a clear separation between the model and the algorithm used to solve it, and 3) they usually provide some measure of uncertainty about the estimation.

On the other hand, adaptive filtering is a standard approach in estimation problems when the input is received as a stream of data that is potentially non-stationary. This approach is widely understood and applied to several problems such as echo cancellation \cite{gilloire1992adaptive}, noise cancellation \cite{nelson1991active}, and channel equalization \cite{falconer2002frequency}.

Although these two approaches share some underlying relations, there are very few connections in the literature. The first important attempt in the signal processing community to relate these two fields was the connection between a linear Gaussian state-space model (i.e. Kalman filter) and the RLS filter, by Sayed and Kailath \cite{sayed1994state} and then by Haykin \emph{et al.} \cite{haykin1997adaptive}. The RLS adaptive filtering algorithm emerges naturally when one defines a particular state-space model (SSM) and then performs exact inference in that model. This approach was later exploited in \cite{van2012kernel} to design a kernel RLS algorithm based on Gaussian processes.

A first attempt to approximate the LMS filter from a probabilistic perspective was presented in \cite{park2014probabilistic}, focusing on a kernel-based implementation. The algorithm of \cite{park2014probabilistic} makes use of a Maximum a Posteriori (MAP) estimate as an approximation for the predictive step. However, this approximation does not preserve the estimate of the uncertainty in each step, therefore degrading the performance of the algorithm.

In this work, we provide a similar connection between state-space models and least-mean-squares (LMS). Our approach is based on approximating the posterior distribution with an isotropic Gaussian distribution. We show how the computation of this approximated posterior leads to a linear-complexity algorithm, comparable to the standard LMS. Similar approaches have already been developed for a variety of problems such as channel equalization using recurrent RBF neural networks \cite{cid1994recurrent}, or Bayesian forecasting \cite{harrison1999bayesian}. Here, we show the usefulness of this probabilistic approach for adaptive filtering.

The probabilistic perspective we adopt throughout this work presents two main advantages. Firstly, a novel LMS algorithm with adaptable step size emerges naturally with this approach, making it suitable for both stationary and non-stationary environments. The proposed algorithm has less free parameters than previous LMS algorithms with variable step size \cite{kwong1992variable,aboulnasr1997robust,shin2004variable}, and its parameters are easier to be tuned w.r.t. these algorithms and standard LMS. Secondly, the use of a probabilistic model provides us with an estimate of the error variance, which is useful in many applications.

Experiments with simulated and real data show the advantages of the presented approach with respect to previous works. However, we remark that the main contribution of this paper is that it opens the door to introduce more Bayesian machine learning techniques, such as variational inference and Monte Carlo sampling methods \cite{barber2012bayesian}, to adaptive filtering.\\

\section{Probabilistic Model}

Throughout this work, we assume the observation model to be linear-Gaussian with the following distribution,

\begin{equation}
p(y_k|\wb_k) = \mathcal{N}(y_k;{\bf x}_k^T \wb_k , \sigma_n^2),
\label{eq:mess_eq}
\end{equation}
where  $\sigma_n^2$ is the variance of the observation noise, ${\bf x}_k$ is the regression vector and $\wb_k$ is the parameter vector to be sequentially estimated, both $M$-dimensional column vectors.

In a non-stationary scenario, $\wb_k$ follows a dynamic process. In particular, we consider a diffusion process (random-walk model) with variance $\sigma_d^2$ for this parameter vector:

\begin{equation}
p(\wb_k|\wb_{k-1})= \mathcal{N}(\wb_k;\wb_{k-1}, \sigma_d^2 {\bf I}),
\label{eq:trans_eq}
\end{equation}
where $\bf I$ denotes the identity matrix. In order to initiate the recursion, we assume the following prior distribution on $\wb_k$

\begin{equation}
p(\wb_0)= \mathcal{N}(\wb_0;0, \sigma_d^2{\bf I}).\nonumber
\end{equation}

\section{Exact inference in this model: Revisiting the RLS filter}

Given the described probabilistic SSM, we would like to infer the posterior probability distribution $p(\wb_k|y_{1:k})$.
Since all involved distributions are Gaussian, one can perform exact inference, leveraging the probability rules in a straightforward manner. The resulting probability distribution is
\begin{equation}
p(\wb_k|y_{1:k}) =  \mathcal{N}(\wb_k;{\bf\mub}_{k}, \Sigmab_{k}), \nonumber
\end{equation}
in which the mean vector ${\bf\mub}_{k}$ is given by
\begin{equation}
{\bf\mub}_k = {\bf\mub}_{k-1} + {\bf K}_k (y_k - {\bf x}_k^T {\bf\mub}_{k-1}){\bf x}_k, \nonumber
\end{equation}
where we have introduced the auxiliary variable
\begin{equation}
{\bf K}_k = \frac{ \left(\Sigmab_{k-1} + \sigma_d^2 {\bf I}\right)}{{\bf x}_k^T  \left(\Sigmab_{k-1} + \sigma_d^2 {\bf I}\right)  {\bf x}_k + \sigma_n^2}, \nonumber
\end{equation}
and the covariance matrix $\Sigmab_k$ is obtained as
%
%
\begin{equation}
\Sigmab_k = \left( {\bf I} -  {\bf K}_k{\bf x}_k {\bf x}_k^T \right) ( \Sigmab_{k-1} +\sigma_d^2), \nonumber
\end{equation}
%
Note that the mode of $p(\wb_k|y_{1:k})$, i.e. the maximum-a-posteriori estimate (MAP), coincides with the RLS adaptive rule
\begin{equation}
{\wb}_k^{(RLS)} = {\wb}_{k-1}^{(RLS)} + {\bf K}_k (y_k - {\bf x}_k^T {\wb}_{k-1}^{(RLS)}){\bf x}_k .
\label{eq:prob_rls}
\end{equation}
This rule is similar to the one introduced in \cite{haykin1997adaptive}.

Finally, note that the covariance matrix $\Sigmab_k$ is a measure of the uncertainty of the estimate $\wb_k$ conditioned on the observed data $y_{1:k}$. Nevertheless, for many applications a single scalar summarizing the variance of the estimate could prove to be sufficiently useful. In the next section, we show how such a scalar is obtained naturally when $p(\wb_k|y_{1:k})$ is approximated with an isotropic Gaussian distribution. We also show that this approximation leads to an LMS-like estimation.

\section{Approximating the posterior distribution: LMS filter }

The proposed approach consists in approximating the posterior distribution $p(\wb_k|y_{1:k})$, in general a multivariate Gaussian distribution with a full covariance matrix, by an isotropic spherical Gaussian distribution 

\begin{equation}
\label{eq:aprox_post}
\hat{p}(\wb_{k}|y_{1:k})=\mathcal{N}(\wb_{k};{\bf \hat{\mub}}_{k}, \hat{\sigma}_{k}^2 {\bf I} ).
\end{equation}

In order to estimate the mean and covariance of the approximate distribution $\hat{p}(\wb_{k}|y_{1:k})$, we propose to select those that minimize the Kullback-Leibler divergence with respect to the original distribution, i.e., 

\begin{equation}
\{\hat{\mub}_k,\hat{\sigma}_k\}=\arg \displaystyle{  \min_{\hat{\mub}_k,\hat{\sigma}_k}} \{ D_{KL}\left(p(\wb_{k}|y_{1:k}))\| \hat{p}(\wb_{k}|y_{1:k})\right) \}. \nonumber
\end{equation}

The derivation of the corresponding minimization problem can be found in Appendix A. In particular, the optimal mean and the covariance are found as
\begin{equation}
{\hat{\mub}}_{k} = {\mub}_{k};~~~~~~ \hat{\sigma}_{k}^2 = \frac{{\sf Tr}\{ \Sigmab_k\} }{M}.
\label{eq:sigma_hat}
\end{equation}
%

%

We now show that by using \eqref{eq:aprox_post} in the recursive predictive and filtering expressions we obtain an LMS-like adaptive rule. First, let us assume that we have an approximate posterior distribution at $k-1$, $\hat{p}(\wb_{k-1}|y_{1:k-1}) =  \mathcal{N}(\wb_{k-1};\hat{\bf\mub}_{k-1}, \hat{\sigma}_{k-1}^2 {\bf I} )$. Since all involved distributions are Gaussian, the predictive distribution
is obtained as %
\begin{eqnarray}
\hat{p}(\wb_k|y_{1:k-1}) &=& \int p(\wb_k|\wb_{k-1}) \hat{p}(\wb_{k-1}|y_{1:k-1}) d\wb_{k-1} \nonumber\\
&=& \mathcal{N}(\wb_k;{\bf\mub}_{k|k-1}, \Sigmab_{k|k-1}), 
\label{eq:approx_pred}
\end{eqnarray}
where the mean vector and covariance matrix are given by
\begin{eqnarray}
\hat{\bf\mub}_{k|k-1} &=& \hat{\bf\mub}_{k-1} \nonumber \\
\hat{\Sigmab}_{k|k-1} &=& (\hat{\sigma}_{k-1}^2 + \sigma_d^2 ){\bf I}\nonumber.
\end{eqnarray}

From \eqref{eq:approx_pred}, the posterior distribution at time $k$ can be computed using Bayes' Theorem and standard Gaussian manipulations (see for instance \cite[Ch. 4]{murphy2012machine}). Then, we approximate the posterior $p(\wb_k|y_{1:k})$ with an isotropic Gaussian,
%
%
\begin{equation}
\hat{p}(\wb_k|y_{1:k}) =  \mathcal{N}(\wb_k ; {\hat{\mub}}_{k}, \hat{\sigma}_k^2 {\bf I} ),\nonumber
\end{equation}
where 
\begin{eqnarray}
{\hat{\mub}}_{k} &= & {\hat{\mub}}_{k-1}+ \frac{ (\hat{\sigma}_{k-1}^2+ \sigma_d^2)  }{(\hat{\sigma}_{k-1}^2+ \sigma_d^2)  \|{\bf x}_k\|^2 + \sigma_n^2} (y_k - {\bf x}_k^T {\hat{\mub}}_{k-1}){\bf x}_k  \nonumber  \\
&=& {\hat{\mub}}_{k-1}+ \eta_k (y_k - {\bf x}_k^T {\hat{\mub}}_{k-1}){\bf x}_k . 
\label{eq:prob_lms}
\end{eqnarray}
Note that, instead of a gain matrix ${\bf K}_k$ as in Eq.~\eqref{eq:prob_rls}, we now have a scalar gain $\eta_k$ that operates as a variable step size.

Finally, to obtain the posterior variance, which is our measure of uncertainty, we apply \eqref{eq:sigma_hat} and the trick ${\sf Tr}\{{\bf x}_k{\bf x}_k^T\}= {\bf x}_k^T{\bf x}_k= \|{\bf x}_k \|^2$,

\begin{eqnarray}
\hat{\sigma}_k^2 &=& \frac{{\sf Tr}(\Sigmab_k)}{M} \\
&=& \frac{1}{M}{\sf Tr}\left\{ \left( {\bf I} -  \eta_k {\bf x}_k {\bf x}_k^T \right) (\hat{\sigma}_{k-1}^2 +\sigma_d^2)\right\} \\
&=& \left(1 - \frac{\eta_k \|{\bf x}_k\|^2}{M}\right)(\hat{\sigma}_{k-1}^2 +\sigma_d^2).
\label{eq:sig_k}
\end{eqnarray}
If MAP estimation is performed, we obtain  an adaptable step-size LMS estimation

\begin{equation}
\wb_{k}^{(LMS)} = \wb_{k-1}^{(LMS)} + \eta_k (y_k - {\bf x}_k^T \wb_{k-1}^{(LMS)}){\bf x}_k, 	
\label{eq:lms}
\end{equation}
with
\begin{equation}
\eta_k = \frac{ (\hat{\sigma}_{k-1}^2+ \sigma_d^2)  }{(\hat{\sigma}_{k-1}^2+ \sigma_d^2)  \|{\bf x}_k\|^2 + \sigma_n^2}.\nonumber
\end{equation}
At this point, several interesting remarks can be made:

\begin{itemize}

\item The adaptive rule \eqref{eq:lms} has linear complexity since it does not require us to compute the full matrix $\Sigmab_k$.

\item For a stationary model, we have $\sigma_d^2=0$ in \eqref{eq:prob_lms} and \eqref{eq:sig_k}. In this case, the algorithm remains valid and both the step size and the error variance, $\hat{\sigma}_{k}$, vanish over time $k$. 

\item Finally, the proposed adaptable step-size LMS has only two parameters, $\sigma_d^2$ and $\sigma_n^2$, (and only one, $\sigma_n^2$, in stationary scenarios) in contrast to other variable step-size algorithms \cite{kwong1992variable,aboulnasr1997robust,shin2004variable}. More interestingly, both $\sigma_d^2$ and $\sigma_n^2$ have a clear underlying physical meaning, and they can be estimated in many cases. We will comment more about this in the next section. 
\end{itemize}

\section{Experiments}
\label{sec:experiments}

We evaluate the performance of the proposed algorithm in both stationary and tracking experiments. In the first experiment, we estimate a fixed vector $\wb^{o}$ of dimension $M=50$. The entries of the vector are independently and uniformly chosen in the range $[-1,1]$. Then, the vector is normalized so that $\|\wb^o\|=1$. Regressors $\boldsymbol{x}_{k}$ are zero-mean Gaussian vectors with identity covariance matrix. The additive noise variance is such that the SNR is $20$ dB. We compare our algorithm with standard RLS and three other LMS-based algorithms: LMS, NLMS \cite{sayed2008adaptive}, VSS-LMS \cite{shin2004variable}.\footnote{The used parameters for each algorithm are: for RLS $\lambda=1$, $\epsilon^{-1}=0.01$; for LMS $\mu=0.01$; for NLMS $\mu=0.5$; and for VSS-LMS $\mu_{max}=1$, $\alpha=0.95$, $C=1e-4$.} The probabilistic LMS algorithm in \cite{park2014probabilistic} is not simulated because it is not suitable for stationary environments.

\begin{figure}[htb]
\centering
\begin{minipage}[b]{\linewidth}
  \centering
  \centerline{\includegraphics[width=\textwidth]{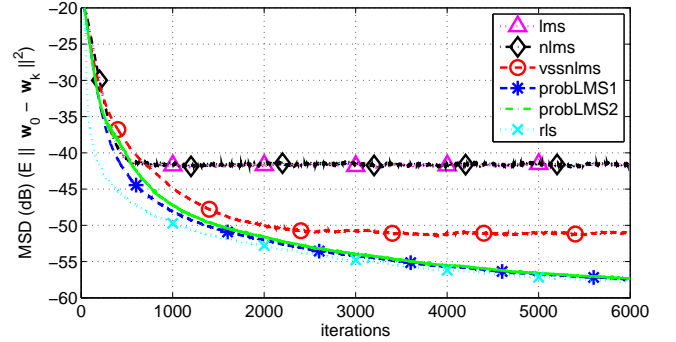}}
\end{minipage}
\caption{Performance in terms of MSD of probabilistic LMS with both optimal (probLMS1) and suboptimal (probLMS2) compared to LMS, NLMS, VS-LMS, and RLS.}
\label{fig:msd_statationary}
\end{figure}

In stationary environments, the proposed algorithm has only one parameter, $\sigma^2_n$. We simulate both the scenario where we have perfectly knowledge of the amount of noise (probLMS1) and the case where the value $\sigma^2_n$ is $100$ times smaller than the actual value (probLMS2). The Mean-Square Deviation (${\sf MSD} = \EE \| \wb_0 - \wb_k \|^2$), averaged out over $50$ independent simulations, is presented in Fig. \ref{fig:msd_statationary}.


The performance of probabilistic LMS is close to RLS (obviously at a much lower computational cost) and largely outperforms previous variable step-size LMS algorithms proposed in the literature. Note that, when the model is stationary, i.e. $\sigma^2_d=0$ in \eqref{eq:trans_eq},  both the uncertainty $\hat{\sigma}^2_k$, and the adaptive step size $\eta_k$, vanish over time. This implies that the error tends to zero when $k$ goes to infinity. Fig. \ref{fig:msd_statationary} also shows that the proposed approach is not very sensitive to a bad choice of its only parameter, as demonstrated by the good results of probLMS2, which uses a $\sigma^2_n$ that is $100$ times smaller than the optimal value.

\begin{figure}[htb]
\centering
\begin{minipage}[b]{\linewidth}
  \centering
  \centerline{\includegraphics[width=\textwidth]{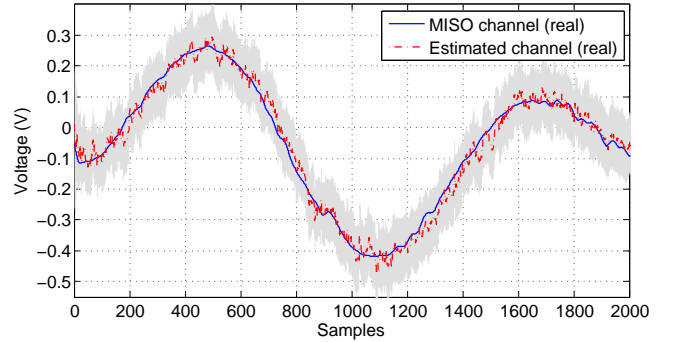}}
\end{minipage}
\caption{Real part of one coefficient of the measured and estimated channel in experiment two. The shaded area represents two standard deviations from the prediction {(the mean of the posterior distribution)}.}
\label{fig_2}
\end{figure}

\begin{table}[ht]
\begin{footnotesize}
\setlength{\tabcolsep}{2pt}
\def\marginwidth{1.5mm}
\begin{center}
\begin{tabular}{|l@{\hspace{\marginwidth}}|c@{\hspace{\marginwidth}}|c@{\hspace{\marginwidth}}|c@{\hspace{\marginwidth}}|c@{\hspace{\marginwidth}}|c@{\hspace{\marginwidth}}|c@{\hspace{\marginwidth}}|}
\hline
Method &  LMS &  NLMS & LMS-2013 & VSSNLMS & probLMS & RLS \\
\hline
\hline
MSD (dB) &-28.45 &-21.07 &-14.36 &-26.90 &-28.36 &-25.97\\ 
\hline                                                                     
\end{tabular}
\end{center}
\caption{Steady-state MSD of the different algorithms for the tracking of a real MISO channel.}
\label{tab:table_MSD}
\end{footnotesize}

\end{table}
\newpage
In a second experiment, we test the tracking capabilities of the proposed algorithm with {real} data of a wireless MISO channel acquired in a realistic indoor scenario. More details on the setup can be found in \cite{gutierrez2011frequency}. Fig. \ref{fig_2} shows the real part of one of the channels, and the estimate of the proposed algorithm. The shaded area represents the estimated uncertainty for each prediction, i.e. $\hat{\mu}_k\pm2\hat{\sigma}_k$. Since the experimental setup does not allow us to obtain the optimal values for the parameters, we fix these parameters to their values that optimize the steady-state mean square deviation (MSD). \hbox{Table \ref{tab:table_MSD}} shows this steady-state MSD of the estimate of the MISO channel with different methods. As can be seen, the best tracking performance is obtained by standard LMS and the proposed method. 


\section{Conclusions and Opened Extensions}
\label{sec:conclusions}

{We have presented a probabilistic interpretation of the least-mean-square filter. The resulting algorithm is an adaptable step-size LMS that performs well both in stationary and tracking scenarios. Moreover, it has fewer free parameters than previous approaches and these parameters have a clear physical meaning. Finally, as stated in the introduction, one of the advantages of having a probabilistic model is that it is easily extensible:}

\begin{itemize}
\item If, instead of using an isotropic Gaussian distribution in the approximation, we used a Gaussian with diagonal covariance matrix, we would obtain a similar algorithm with different step sizes and measures of uncertainty, for each component of $\wb_k$. Although this model can be more descriptive, it needs more parameters to be tuned, and the parallelism with LMS vanishes.
\item Similarly, if we substitute the transition model of \eqref{eq:trans_eq} by an Ornstein-Uhlenbeck process, 

\begin{equation}
p(\wb_k|\wb_{k-1})= \mathcal{N}(\wb_k;\lambda \wb_{k-1}, \sigma_d^2), \nonumber
\label{eq:trans_eq_lambda}
\end{equation}
a similar algorithm is obtained but with a forgetting factor $\lambda$ multiplying $\wb_{k-1}^{(LMS)}$ in \eqref{eq:lms}. This algorithm may have improved performance under such a kind of autoregresive dynamics of $\wb_{k}$, though, again, the connection with standard LMS becomes dimmer.

\item As in \cite{park2014probabilistic}, the measurement model \eqref{eq:mess_eq} can be changed to obtain similar adaptive algorithms for classification, ordinal regression, and Dirichlet regression for compositional data. 

\item A similar approximation technique could be applied to more complex dynamical models, i.e. switching dynamical models \cite{barber2010graphical}. The derivation of efficient adaptive algorithms that explicitly take into account a switch in the dynamics of the parameters of interest is a non-trivial and open problem, though the proposed approach could be useful.

\item Finally, like standard LMS, this algorithm can be kernelized for its application in estimation under non-linear scenarios.

\end{itemize}

\begin{appendices}

\section{KL divergence between a general gaussian distribution and an isotropic gaussian}
\label{sec:kl}

 We want to approximate  $p_{\xb_1}(x) = \mathcal{N}(\xb; \mub_1,\Sigmab_1)$ by $p_{\xb_2}(\xb) = \mathcal{N}(\xb; \mub_2,\sigma_2^2 {\bf I})$. In order to do so, we have to compute the parameters of $p_{\xb_2}(\xb)$, $\mub_2$ and $\sigma_2^2$, that minimize the following Kullback-Leibler divergence,

\begin{eqnarray}
D_{KL}(p_{\xb_1}\| p_{\xb_2}) &=&\int_{-\infty}^{\infty} p_{\xb_1}(\xb) \ln{\frac{p_{\xb_1}(\xb)}{p_{\xb_2}(\xb)}}d\xb \nonumber  \\
&= &  \frac{1}{2} \{ -M + {\sf Tr}(\sigma_2^{-2} {\bf I}\cdot \Sigmab_1^{-1})  \nonumber \\
  & &  + (\mub_2 - \mub_1 )^T \sigma^{-2}_2{\bf I} (\mub_2 - \mub_1 )  \nonumber \\
 & &   +  \ln \frac{{\sigma_2^2}^M}{\det\Sigmab_1} \}.  
\label{eq:divergence}
\end{eqnarray}
Using symmetry arguments, we obtain 
\begin{equation}
\mub_2^{*} =\arg \displaystyle{  \min_{\mub_2}} \{ D_{KL}(p_{\xb_1}\| p_{\xb_2}) \} = \mub_1.
\end{equation}
Then, \eqref{eq:divergence} gets simplified into 

\begin{eqnarray}
D_{KL}(p_{\xb_1}\| p_{\xb_2}) = \frac{1}{2}\lbrace { -M + {\sf Tr}(\frac{\Sigmab_1}{\sigma_2^{2}}) + \ln \frac{\sigma_2^{2M}}{\det\Sigmab_1}}\rbrace.
\end{eqnarray}
The variance $\sigma_2^2$ is computed in order to minimize this Kullback-Leibler divergence as

\begin{eqnarray}
\sigma_2^{2*} &=& \arg\min_{\sigma_2^2} D_{KL}(P_{x_1}\| P_{x_2}) \nonumber \\
 &=& \arg\min_{\sigma_2^2}\{ \sigma_2^{-2}{\sf Tr}\{\Sigmab_1\} + M\ln \sigma_2^{2} \} .
\end{eqnarray}
Deriving and making it equal zero leads to

\begin{equation}
\frac{\partial}{\partial \sigma_2^2} \left[ \frac{{\sf Tr}\{\Sigmab_1\}}{\sigma_2^{2}} + M \ln \sigma_2^{2} \right] = \left. {\frac{M}{\sigma_2^{2}}-\frac{{\sf Tr}\{\Sigmab_1\}}{(\sigma_2^{2})^2}}\right|_{\sigma_2^{2}=\sigma_2^{2*}}\left. =0 \right. .
\nonumber
\end{equation}
Finally, since the divergence has a single extremum in $R_+$,
\begin{equation}
\sigma_2^{2*} = \frac{{\sf Tr}\{\Sigmab_1\}}{M}.
\end{equation}

%
%
%
%
%
%
%
%
%
%
%

\end{appendices}

\vfill
\clearpage

\bibliographystyle{IEEEbib}

\end{document}